# A Two-stage Fine-tuning Strategy for Generalizable Manipulation Skill of Embodied AI

Fang Gao[1†], XueTao Li[1†*], Jun Yu[2], Feng Shaung[1]

*Abstract*—The advent of Chat-GPT has led to a surge of interest in Embodied AI. However, many existing Embodied AI models heavily rely on massive interactions with training environments, which may not be practical in real-world situations. To this end, the Maniskill2 has introduced a full-physics simulation benchmark for manipulating various 3D objects. This benchmark enables agents to be trained using diverse datasets of demonstrations and evaluates their ability to generalize to unseen scenarios in testing environments. In this paper, we propose a novel two-stage fine-tuning strategy that aims to further enhance the generalization capability of our model based on the Maniskill2 benchmark. Through extensive experiments, we demonstrate the effectiveness of our approach by achieving the 1st prize in all three tracks of the ManiSkill2 Challenge. Our findings highlight the potential of our method to improve the generalization abilities of Embodied AI models and pave the way for their practical applications in real-world scenarios. All codes and models of our solution is available at https://github.com/xtli12/GXU-LIPE.git

*Keywords-component; Embodied AI; Maniskill2; two-stage fine-tuning*

## I. Introduction

With the rise of Chat-GPT [1], AI (artificial intelligence) has once again sparked a global frenzy. But models like GPT [2], despite being large, do not have a tangible impact on the physical world. On the other hand, Embodied AI goes a step further by incorporating a physical body. It gathers environmental information through sensors and carries out physical actions using mechanical actuators. It can also engage in real-time interactions with humans and the environment through physical entities such as robots.

With the assistance of Embodied AI, robots would be able to automate daily tasks such as household chores. To achieve this goal, however, robots must possess human-like manipulation skills which is extraordinary in the sense that we can manipulate unseen things of the same category once we learn to manipulate them. However, many existing Embodied AI models heavily rely on extensive interactions with training environments, which may not be practical or feasible in real-world scenarios. To this end, SAPIEN ManiSkill [3] has proposed a full-physics simulation benchmark for manipulating a variety of 3D objects. This benchmark allows agents to be trained using a large-scale dataset of demonstrations and assesses their generalization ability in unseen scenarios in testing environments. By leveraging this benchmark, agents can be trained to possess the necessary adaptability and generalization skills required for real-world applications. SAPIEN ManiSkill simulates a panoramic camera in motion and provides 3D deep learning algorithms to train the agent. ManiSkill2 [4], the next version of the SAPIEN ManiSkill, has incorporated a variety of manipulation assignments to resolve the generalizability issue in the training of manipulation skills. Based on ManiSkill2, the ManiSkill2 challenge utilizes a variety of articulated objects in a full-physics simulator to accomplish 20 tasks (see Fig. 1), those tasks can be divided into two main groups: soft-body tasks and rigid-body tasks. More details are as follows:

### A. Soft-body tasks

ManiSkill2 challenge includes 6 soft-body manipulation tasks that call for agents to engage with soft bodies, moving or deforming them to achieve predetermined target states.

- *Fill*: Filling the target beaker with clay from a bucket;
- *Hang*: Hanging a noodle on the target pole;
- *Excavate*: Gathering a specified quantity of clay and elevating it to a specified height;
- *Pour*: Pouring water into the target beaker from a bottle;
- *Pinch*: Deforming plasticine into a target shape by randomly pinching initial shapes;
- *Write*: Writing a target character on clay.

### B. Rigid-body tasks

ManiSkill2 challenge consists of 14 rigid-body manipulation tasks that require agents to interact with rigid objects to accomplish day-to-day tasks. Those tasks include 4 tasks from SAPIEN ManiSkill (*PushChair*, *MoveBucket*, *OpenCabinetDoor*, and *OpenCabinetDrawer*) and 10 novel tasks as follows.

- *StackCube:* Lifting a cube and stacking it on top of another cube;
- *PickSingleYCB*: Selecting and positioning a single YCB object [5];
- *PickSingleEGAD*: Selecting and positioning a single EGAD object [6];
- *PickClutterYCB*: Similar to *PickSingleYCB,* but multiple objects are present in the scene;

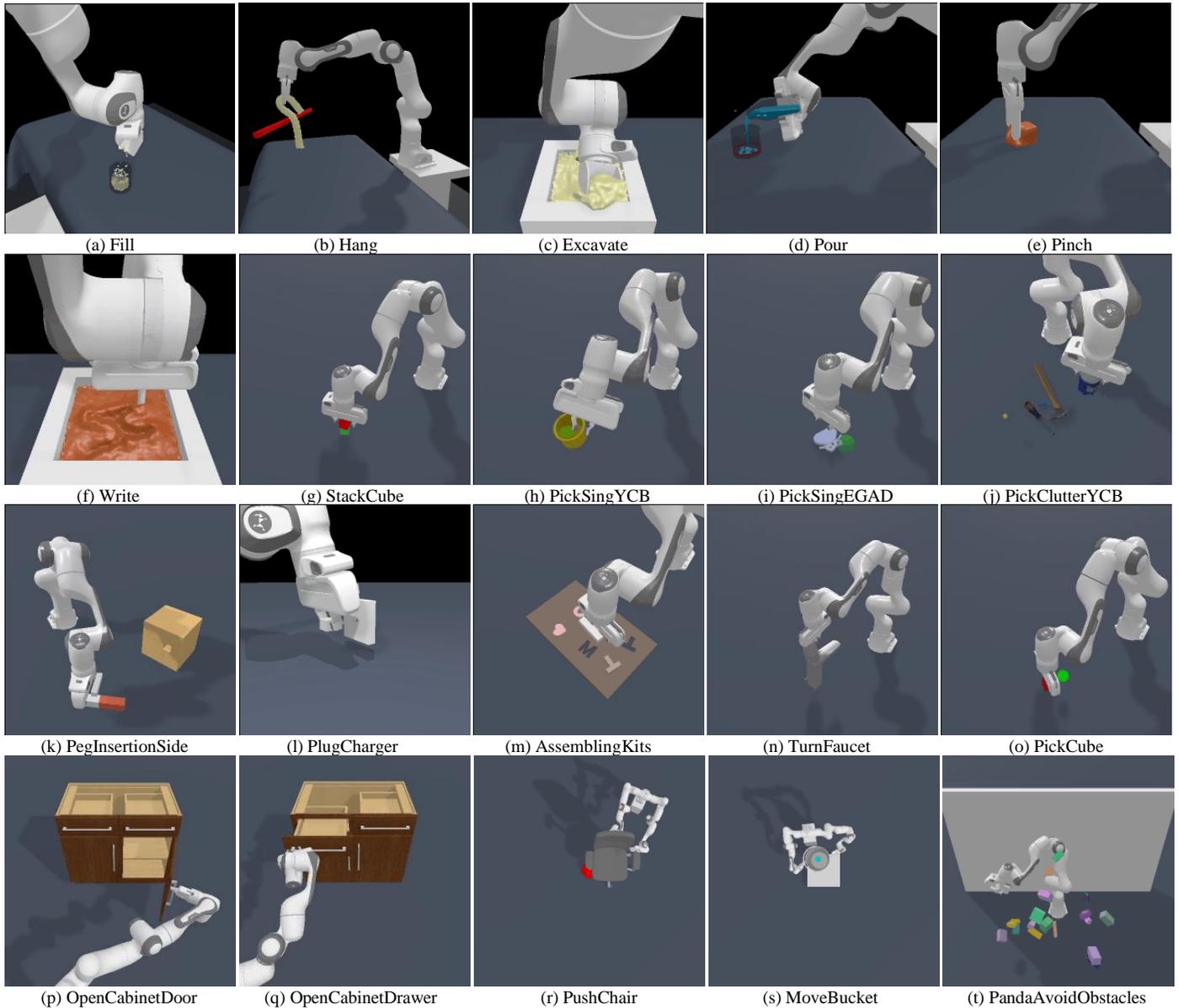
Fig. 1. Tasks in Maniskill2 challenge

- *PegInsertionSide*: A MetaWorld-inspired [7] peg-in-hole construction activity;
- *PlugCharger*: Inspired by RLBench [8], requiring an agent to pick up and plug a charger into a vertical socket;
- *AssemblingKits*: Inspired by Transporter Networks [9], involving the ability of an agent to pick up and insert a form into its slot on a board with 5 slots;
- *TurnFaucet*: Requiring an agent to turn on faucets of different geometries and topologies using a stationary arm;
- *PickCube*: Picking a cube in a target position;
- *PandaAvoidObstacles*: Evaluating the ability of a stationary limb to navigate through a dense array of obstacles in space while actively perceiving the environment.

In this paper, we apply PointNet [10] to extract the point cloud features, and then use Imitation or Reinforcement Learning algorithms [11] to determine the direction and the moving distance of the agent. For soft-body tasks, we utilize Behavior Cloning (BC) [12], while for rigid-body tasks, we utilize Proximal Policy Gradient (PPO) [13]. Specifically, we propose a two-stage fine-tuning strategy to further explore the potential capacity of the model. The 1st rank in each of the three ManiSkill2 Challenge tracks has demonstrated the efficacy of our method.

II. METHOD

Within the framework of Maniskill2, we have implemented different controllers and policies tailored to various tasks.

TABLE I. SUCCESS RATE OF DIFFERENT SCALES

| Row | Batch size | Samples | PickSingleEGAD | |
|---|---|---|---|---|
| | | | Train | Test |
| 1 | 330 | 20000 | 0.65 | 0.60 |
| 2 | 330 x 0.9 | 20000 x 1 | 0.71 | 0.59 |
| 3 | 330 x 0.8 | 20000 x 1 | 0.57 | 0.49 |
| 4 | 330 x 0.7 | 20000 x 1 | 0.67 | 0.59 |
| **5** | **330 x 0.9** | **20000 x 0.875** | **0.72** | **0.67** |
| 6 | 330 x 0.8 | 20000 x 0.875 | 0.65 | 0.59 |
| 7 | 330 x 0.7 | 20000 x 0.875 | 0.66 | 0.60 |
| 8 | 330 x 0.9 | 20000 x 0.75 | 0.65 | 0.58 |
| 9 | 330 x 0.8 | 20000 x 0.75 | 0.66 | 0.55 |
| 10 | 330 x 0.7 | 20000 x 0.75 | 0.64 | 0.54 |

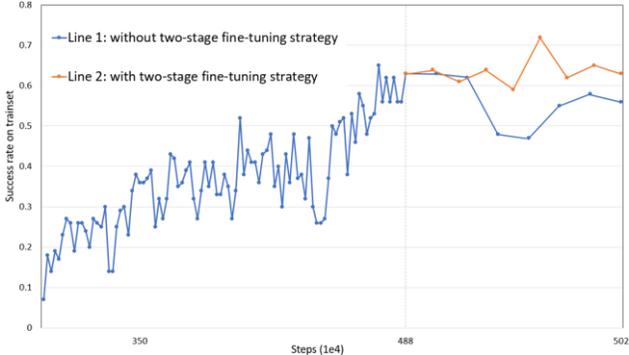

Fig. 2. The trend line of success rate with *PickSingleEGAD* task *(488 denotes the highest score checkpoint of Line 1)*

Controllers serve as a bridge between policies and robots, where policies generate actions. These actions are then translated by controllers into control signals that can be utilized by the joints of robot for movement. The details of the implements are as follows:

*A. Rigid-body tasks*

In rigid-body tasks, the *base-pd-joint-vel-arm-pd-joint-vel* controller is used for *PushChair*, *MoveBucket*, and *OpenCabinetDoor* tasks; the *base-pd-joint-vel-arm-pd-ee-delta-pose* controller is used for *OpenCabinetDrawer* task; the *pd-ee-delta-pose* controller is used for all other tasks. Firstly, we extract the point cloud features by PointNet. These features are then fed into PPO algorithms to generate an action plan. Subsequently, the controller converts these actions to control signals that drive the agent to complete the task.

During the initial stage of fine-tuning, we achieve the highest score on the test set. However, as the training process continued, we observe a decline in the success rate (see Line 1 in Fig. 2), indicating potential overfitting of the model to the specific task. To address this issue and further explore the potential capacity of the model, we introduce a two-stage fine-tuning strategy.

In the second stage of our approach, we resume the training process from the highest score checkpoint obtained in the initial stage. However, in addition to resuming training, we make two important adjustments: reducing the batch size and the number of samples in each step. This reduction encourages the model to pay more attention to smaller volumes of information. However, it is important to note that setting a smaller batch size and fewer samples in each step can introduce more noise into the training process [14], as it may extract more irrelevant information. Nonetheless, this adjustment helps mitigate the overfitting issue to some extent.

Through experimentation, we explore different strategies for decreasing the above-mentioned two parameters and identify the optimal scale that leads to higher scores. By implementing this two-stage fine-tuning strategy, we tap into the potential capacity of the model and improve its performance on the given tasks.

*B. Soft-body tasks*

In soft-body tasks, we use the *pd-joint-delta-pos* controller. The overall policy bears similarities to that in rigid-body tasks, with the primary difference being the substitution of PPO algorithms with BC algorithms. This change is specific to soft-body tasks. To improve performance and attain higher scores in soft-body tasks, we also incorporate our two-stage fine-tuning strategy. By leveraging BC algorithms and implementing the two-stage fine-tuning strategy, we aim to enhance the effectiveness of the model in handling soft-body tasks and achieve improved results.

III. EXPERIMENTS

*A. Two-stage fine-tuning strategy*

In our pursuit of determining the optimal scale of the batch size and the number of samples in each step of the two-stage fine-tuning strategy, we conducted a series of experiments specifically focused on the *PickSingleEGAD* task. We tested various scales, including 0.7, 0.8, and 0.9 for the batch size, and 1, 0.875, and 0.75 for the number of samples in each step.

The results of these experiments, as shown in Table I, indicate that the scale of 0.9 for the batch size and 0.875 for the number of samples yielded better performance. To further illustrate the effectiveness of our two-stage fine-tuning strategy, we visualized the trend line of the success rate with and without our strategy in Fig. 2. Applying the scale of 0.875 for the batch size and 0.9 for the number of samples, as shown in Fig. 2, allowed the model to effectively process the training data and mitigate the overfitting issue to some extent.

The improved performance on different tasks, as seen in Table II, after implementing the two-stage fine-tuning strategy, further demonstrates the efficacy of our method. The optimized configuration enabled the model to extract more meaningful information from the data, leading to enhanced performance and generalization across various tasks.

*B. Results on the leaderboard of the Mainskill2023 Challenge*

There are three tracks for the Mainskill2023 Challenge:
- Imitation/Reinforcement Learning (Rigid Body): which is for solutions that utilize IL and/or RL to solve rigid body environments;
- No Restriction (Rigid Body): an open track that permits any kind of solution to solve rigid body environments;

TABLE II. THE IMPROVED PERFORMANCE AFTER APPLYING THE TWO-STAGE FINE-TUNING STRATEGY ON DIFFERENT TASKS

| With or without two-stage fine tuning strategy | StackCube | | PickSingle-YCB | | PickSingle-EGAD | | PlugCharger | | TurnFaucet | | OpenCabinet-Drawer | | OpenCabinet-Door | | PushChair | | MoveBucket | |
|---|---|---|---|---|---|---|---|---|---|---|---|---|---|---|---|---|---|---|
| | *Train* | *Test* | *Train* | *Test* | *Train* | *Test* | *Train* | *Test* | *Train* | *Test* | *Train* | *Test* | *Train* | *Test* | *Train* | *Test* | *Train* | *Test* |
| With | **1** | **0.99** | **0.66** | **0.55** | **0.69** | **0.67** | **0.04** | **0.03** | **0.28** | **0.23** | **0.15** | **0.16** | **0.59** | **0.24** | 0.04 | 0.08 | 0.02 | 0.08 |
| Without | 0.96 | 0.99 | 0.51 | 0.45 | 0.54 | 0.49 | 0.02 | 0.01 | 0.20 | 0.14 | 0.12 | 0.10 | 0.48 | 0.08 | 0.04 | 0.08 | 0.02 | 0.08 |

TABLE III. THE LEADERBOARD OF IMITATION/REINFORCEMENT LEARNING (TEST/RIGID BODY)

| Team Name | Average | Stack-Cube | Plug-Charger | Peg-Insertion-Side | Turn-Faucet *(test)* | Pick-Clutter *(test)* | Push-Chair *(test)* | Move-Bucket *(test)* | Pick-Single-YCB *(test)* | Pick-Single-EGAD *(test)* | Assem-bling-Kits *(test)* | Open-Cabinet-Door *(test)* | Open-Cabinet-Drawer *(test)* |
|---|---|---|---|---|---|---|---|---|---|---|---|---|---|
| **GXU-LIPE (Our)** | **0.2499** | 0.99 | **0.03** | 0 | **0.235** | 0 | **0.04** | **0.08** | **0.55** | 0.67 | 0 | **0.24** | **0.16** |
| ChenBao | 0.2085 | 1 | 0.02 | 0 | 0.084 | 0 | 0 | 0 | 0.55 | 0.84 | 0 | 0 | 0 |
| dee | 0.1763 | 0.96 | 0.01 | 0 | 0.026 | 0 | 0 | 0 | 0.46 | 0.66 | 0 | 0 | 0 |
| Joy-3D | 0.0817 | 0.98 | 0 | 0 | 0 | 0 | 0 | 0 | 0 | 0 | 0 | 0 | 0 |
| ApexRL | 0.0408 | 0.49 | 0 | 0 | 0 | 0 | 0 | 0 | 0 | 0 | 0 | 0 | 0 |
| artefacts | 0 | 0 | 0 | 0 | 0 | 0 | 0 | 0 | 0 | 0 | 0 | 0 | 0 |

TABLE IV. THE LEADERBOARD OF NO RESTRICTION (TEST/RIGID BODY)

| Team Name | Average | Stack-Cube | Plug-Charger | Peg-Insertion-Side | Turn-Faucet *(test)* | Pick-Clutter *(test)* | Push-Chair *(test)* | Move-Bucket *(test)* | Pick-Single-YCB *(test)* | Pick-Single-EGAD *(test)* | Assem-bling-Kits *(test)* | Open-Cabinet-Door *(test)* | Open-Cabinet-Drawer *(test)* |
|---|---|---|---|---|---|---|---|---|---|---|---|---|---|
| **GXU-LIPE (Our)** | **0.2499** | 0.99 | **0.03** | 0 | **0.235** | 0 | **0.08** | **0.08** | **0.55** | 0.67 | 0 | **0.24** | **0.12** |
| ChenBao | 0.2085 | 1 | 0.02 | 0 | 0.084 | 0 | 0 | 0 | 0.55 | 0.84 | 0 | 0 | 0 |
| dee | 0.1634 | 0.96 | 0.01 | 0 | 0.026 | 0 | 0 | 0 | 0.305 | 0.66 | 0 | 0 | 0 |

TABLE V. THE LEADERBOARD OF NO RESTRICTION (TEST/SOFT BODY)

| Team Name | Average | Fill | Excavate | Pour | Hang | Write *(test)* | Pinch *(test)* |
|---|---|---|---|---|---|---|---|
| **GXU-LIPE (Our)** | **0.3967** | **0.9** | **0.4** | **0.18** | **0.9** | 0 | 0 |
| ChenBao | 0.1800 | 0.64 | 0 | 0.12 | 0.32 | 0 | 0 |
| dee | 0.1400 | 0.18 | 0.1 | 0 | 0.56 | 0 | 0 |
| Joy-3D | 0 | 0 | 0 | 0 | 0 | 0 | 0 |
| ApexRL | 0 | 0 | 0 | 0 | 0 | 0 | 0 |
| artefacts | 0 | 0 | 0 | 0 | 0 | 0 | 0 |

- No Restriction (Soft Body): an open track that permits any solution to solve soft body environments.

Our method is the 1st of all the three tracks, the result is shown in Table III, Table IV, and Table V, respectively (The leaderboard only evaluates 18 tasks, *PickCube* and *PandaAvoidObstacles* are not included).

## IV. CONCLUSION

In this paper, we utilize the PointNet framework to extract point cloud features. These features serve as input to Imitation or Reinforcement Learning algorithms, enabling us to determine the direction and moving distance of the agent. For soft-body tasks, we employ the BC algorithm, while for rigid-body tasks, we utilize the PPO algorithm. To further explore the potential capacity of model, we propose a two-stage fine-tuning strategy. This strategy has shown remarkable effectiveness, as evidenced by achieving the 1st rank in each of the three ManiSkill2 Challenge tracks.